\documentclass[10pt,journal,compsoc]{IEEEtran}

\usepackage{times}
\usepackage{epsfig}
\usepackage{graphicx}
\usepackage{amsmath}
\usepackage{amssymb}
\usepackage{graphicx}

\newcommand{\tabincell}[2]{\begin{tabular}{@{}#1@{}}#2\end{tabular}}

\ifCLASSOPTIONcompsoc
  \usepackage[nocompress]{cite}
\else
  \usepackage{cite}
\fi
\ifCLASSINFOpdf
\else
\fi
\begin{document}
\bibliographystyle{unsrt}
%
\title{DAFAR: Defending against Adversaries by Feedback-Autoencoder Reconstruction}
%
%
%
%

\author{Haowen~Liu,
        Ping~Yi,~\IEEEmembership{Senior Member,~IEEE},
        Hsiao-Ying~Lin,~\IEEEmembership{Member,~IEEE},
        Jie~Shi,
        and~Weidong~Qiu,~\IEEEmembership{Senior Member,~IEEE}
\IEEEcompsocitemizethanks{
\IEEEcompsocthanksitem H. Liu, P. Yi and W. Qiu are with School of Cyber Science and Engineering, Shanghai Jiao Tong University, No.800 Dongchuan Road, Minhang Area 201100, Shanghai, China.\protect\\
E-mail: haowenliew@outlook.com, \{yiping, qiuwd\}@sjtu.edu.cn
\IEEEcompsocthanksitem H. Lin and J. Shi are with HUAWEI International, Singapore.\protect\\
E-mail: Lin.hsiao.ying@huawei.com, Shi.jie1@huawei.com
}
\thanks{(Corresponding author: Ping Yi.)}
}

%
%

\markboth{Journal of \LaTeX\ Class Files,~Vol.~14, No.~8, August~2015}%
{Haowen \MakeLowercase{\textit{et al.}}: DAFAR: Defending against Adversaries by Feedback-Autoencoder Reconstruction}
%



\IEEEtitleabstractindextext{%
\begin{abstract}
Deep learning has shown impressive performance on challenging perceptual tasks and has been widely used in software to provide intelligent services. However, researchers found deep neural networks vulnerable to adversarial examples. Since then, many methods are proposed to defend against adversaries in inputs, but they are either attack-dependent or shown to be ineffective with new attacks. And most of existing techniques have complicated structures or mechanisms that cause prohibitively high overhead or latency, impractical to apply on real software. We propose DAFAR, a feedback framework that allows deep learning models to detect/purify adversarial examples in high effectiveness and universality, with low area and time overhead. DAFAR has a simple structure, containing a victim model, a plug-in feedback network, and a detector. The key idea is to import the high-level features from the victim model's feature extraction layers into the feedback network to reconstruct the input. This data stream forms a feedback autoencoder. For strong attacks, it transforms the imperceptible attack on the victim model into the obvious reconstruction-error attack on the feedback autoencoder directly, which is much easier to detect; for weak attacks, the reformation process destroys the structure of adversarial examples. Experiments are conducted on MNIST and CIFAR-10 data-sets, showing that DAFAR is effective against popular and arguably most advanced attacks without losing performance on legitimate samples, with high effectiveness and universality across attack methods and parameters.
\end{abstract}

}

\maketitle

\IEEEdisplaynontitleabstractindextext

%
\IEEEpeerreviewmaketitle

\ifCLASSOPTIONcompsoc
\IEEEraisesectionheading{\section{Introduction}\label{sec:introduction}}
\else
\section{Introduction}
\label{sec:introduction}
\fi

%
%
%
%
\IEEEPARstart{D}{eep} learning system plays an increasingly important role in people's everyday life in recent years. However, researchers found deep neural networks (DNNs) to be vulnerable to \emph{adversarial examples} by applying specially crafted perturbations imperceptible to humans on the original samples \cite{szegedy2013intriguing, goodfellow2014explaining, papernot2016limitations, carlini2017towards, madry2017towards, zhang2020generating}. The adversarial examples can cause deep learning models to give a wrong classification, which cast a great threat on modern deep learning systems. With the widespread application of deep learning technology in software, adversarial examples also cause nonnegligible harm to the reliability and dependability of software in real world (e.g., cell-phone camera attack \cite{kurakin2016adversarial}, attack on Autonomous Vehicles \cite{athalye2018synthesizing}, and cyberspace attack \cite{papernot2017practical}).

With advanced adversarial attack methods continuing appearing, researches of defense against adversarial examples also have lots of breakthroughs \cite{meng2017magnet, xu2017feature, gu2014towards, bhagoji2018enhancing, Jia_2019_CVPR, liu2020defending}. However, most defense methods either target specific attacks or were shown to be ineffective with new attacks. Some defense methods only focus on properties of specific attack but ignore common properties of adversarial examples (e.g., adversarial training \cite{andriushchenko2020understanding} and adversary detector \cite{metzen2017detecting, grosse2017statistical}), leading to attack-dependence. Other defense methods with relatively high universality are often easily broken down by strong attack (e.g., Gradient Masking \cite{gu2014towards}, Input Transformation \cite{bhagoji2018enhancing} and MagNet \cite{meng2017magnet}). Furthermore, most of these existing techniques have complicated structures or mechanisms, which adds them prohibitively high area, time or energy overhead, making them impractical to apply on real software.

\begin{figure}[t]
\begin{center}
  \includegraphics[width=1\linewidth, trim=130 60 200 25,clip]{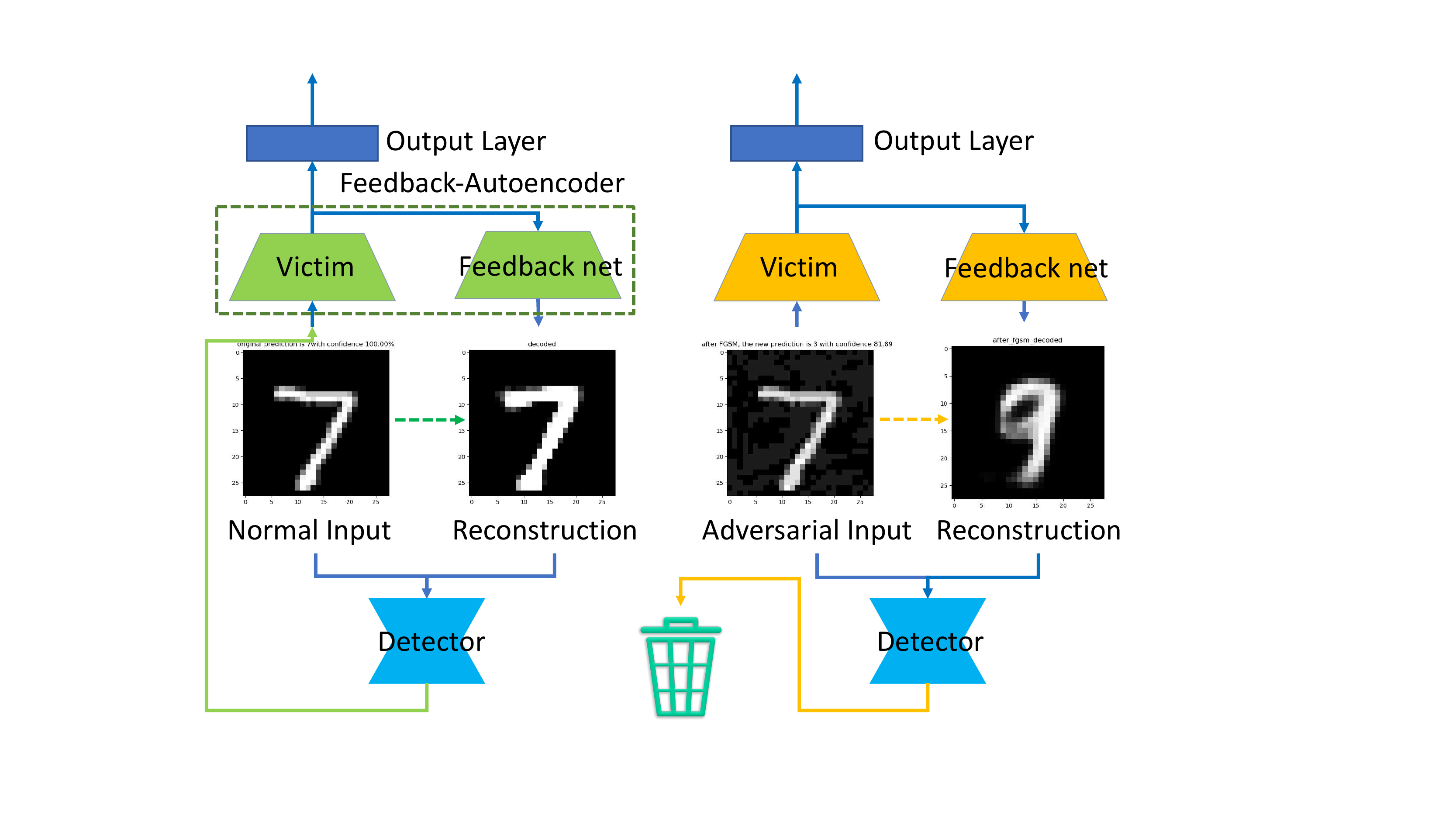}
\end{center}
   \caption{Framework and workflow of DAFAR. The feedback network reconstructs the input from high-level features, and then the detector compares the difference between the input sample and its reconstruction, and tells whether the input is adversarial. The reconstruction of dis-adversarial judgments will be put back into the victim network for classification, and adversarial judgements will be discarded. The left is a normal input case while the right shows an adversarial case.}
\label{fig:framework}
\end{figure}

We propose DAFAR\footnote{DAFAR is the abbreviation for \textbf{D}efending against \textbf{A}dversaries by \textbf{F}eedback-\textbf{A}utoencoder \textbf{R}econstruction.}, shown in Figure \ref{fig:framework}, the first \emph{autogenous hybrid defense} method to defend against adversarial examples. Besides the victim network, DAFAR only contains a feedback network and a detector that can optionally be an anomaly detector. The feature extraction layers of the victim network and the feedback network constitute a feedback autoencoder. Namely, the victim network and the feedback autoencoder share the feature extraction layers, i.e. encoder, and also share the interference introduced by possible adversarial perturbations in high-level feature space. The encoder first extracts high-level features of the input sample, and then pass them to the feedback network. The feedback network reconstruct the sample from these features, and the detector compares the difference between the input sample and its reconstruction, to judge whether the input is adversarial. If so, DAFAR will discard this sample; if not, the reconstruction will be put back into the victim network to be classified instead of the original input. Relying on feedback reconstruction, the \emph{discard data flow} and the \emph{feedback loop data flow} constitute the \emph{autogenous hybrid defense}.

DAFAR has four significant advantages. First, the architecture and mechanism are simple and effortless. DAFAR only contains a feedback network and a detector besides the victim network, and the feedback network can be transformed from the byproduct of victim network's pre-training due to its decoder-basis \cite{hinton2007learning, 10.1007/978-981-15-6353-9_34, kokalj2019mitigation}. And there are only one feedback loop and one error comparison besides the normal classification data flow. Second, it does not modify the victim network, for the feedback network is a plug-in structure, and the detector is an independent part, so it can be used to protect a wide range of neural networks. Third, DAFAR's victim network, feedback network, and the optional anomaly detector are trained on normal samples, without prior knowledge of attack techniques, so DAFAR is attack-independent. Forth, with its autogenous-hybrid-defense strategy, DAFAR achieves considerable defense effectiveness against adversarial examples at whether very low or high attack intensities.


Our main contributions are:
\begin{itemize}
\item We summarize four principles to achieve ideal adversarial example defense effectiveness (Section \ref{motivation}): 1) assume no specific knowledge of attack techniques; 2) for strong attack, detect it; 3) for weak attack, purify it; 4) use a hybrid of detection and purification.

\item Based on the four principles, we design DAFAR, a semi-supervised-trained framework, to defend against adversarial examples in an autogenous-hybrid-defense manner, in high effectiveness and universality (Section \ref{secdesign}).

\item Using some representative DNN models and popular datasets, we demonstrate that DAFAR can effectively defend against the adversarial examples generated by different up-to-date attack techniques with a better performance and simpler mechanism than other representative defense methods (Section \ref{secexp}). 
\end{itemize}

\section{Background}
\subsection{Deep Learning}
Deep learning is a type of machine learning methods based on deep neural networks (DNNs) \cite{LIU201711}, which makes information systems to learn knowledge without explicit programing and extract useful features from raw data. Deep learning models play an increasingly important role in modern life. They are used in image classification \cite{shankar2020hyperparameter, ma2020autonomous}, financial analysis \cite{ozbayoglu2020deep}, disease diagnosis \cite{bi2020computer},  speech recognition \cite{song2020english} and information security \cite{yang2020netflow, al2020ensemble}. 

Several popular network structures are used in daily life and research widely: LeNet \cite{le2013building},  AlexNet \cite{krizhevsky2017imagenet}, VGGNet \cite{simonyan2014very}, GoogLeNet \cite{szegedy2016inception} and ResNet \cite{he2015delving}. Attackers usually generate adversarial examples against these baseline architectures \cite{yuan2019adversarial}.

\subsection{Adversarial Examples}\label{attacks}
Adversarial examples are original clean samples with specially crafted small perturbations, often barely recognizable by humans, but able to misguide the classifier. Since the discovery of adversarial examples for neural networks in \cite{szegedy2013intriguing}, researchers have developed several methods to generate adversarial examples, such as fast gradient sign method (FGSM) \cite{goodfellow2014explaining}, Carlini and Wagner Attacks (CW) \cite{carlini2017towards}, Jacobian-based Saliency Map Attack (JSMA) \cite{papernot2016limitations}, and Projected Gradient Descent (PGD) \cite{madry2017towards}. We will use these four up-to-date attack techniques to craft adversarial examples in our experiments later.

\subsection{Defense Methods}\label{detection}
Adversarial example defense can be categorized into two types: \emph{Only Detection} methods and \emph{Complete Defense} methods \cite{8294186}. And recently some research combined these two ideas, which we call \emph{Hybrid Defense}. In this section we briefly summarize some representative defense methods in these three categories separately.\\

\textbf{Only Detection} is meant to recognize potentially adversarial examples in input to reject them in any further processing. \emph{Only-binary-classifier method} \cite{metzen2017detecting} simply trains a supervised binary classifier on normal and adversarial samples to classify them. \emph{MagNet} \cite{meng2017magnet} contains an autoencoder-based anomaly detector to detect adversaries with large reconstruction errors and a probability-divergence-based detector to handle small ones. \emph{Feature Squeezing} \cite{xu2017feature} detects adversarial examples by comparing a DNN model’s prediction on the original input with that on squeezed inputs. These methods are either attack-dependent or easily cheated by low-intensity attacks, where DAFAR shows a better performance.\\

\textbf{Complete Defense} aims at enabling the victim network to achieve its original goal on the adversarial examples, e.g. a classifier predicting labels of adversarial examples with acceptable accuracy\cite{8294186}. \emph{Adversarial training} techniques \cite{andriushchenko2020understanding} train a more robust model by including adversarial information in training process. \emph{Gradient Masking} techniques \cite{gu2014towards} leverage distillation training techniques \cite{papernot2016distillation} and hide the gradient between the pre-softmax layer (logits) and softmax outputs to defend against gradient-based attacks. \emph{Input Transformation} techniques \cite{bhagoji2018enhancing} like image denoising reduce the model sensitivity to small input changes by transforming the inputs, relieving or changing the input changes, destroying the structure of adversarial example because of its low robustness. \emph{Generative Model Methods} \cite{samangouei2018defense, song2017pixeldefend} find a close output to a given image which does not contain the adversarial changes using a generative model like \emph{Defense-GAN} \cite{samangouei2018defense}. These methods either lose effectiveness in some cases or are easily broken by high-intensity attacks, which is exactly what DAFAR is good at.\\

\textbf{Hybrid Defense} combines the advantages of \emph{Only Detection} and \emph{Complete Defense}, that detection methods are highly effective against high-intensity attacks and purification methods are good at handling low-intensity attacks. To achieve hybrid defense, \emph{MagNet} \cite{meng2017magnet} attaches an autoencoder-based reformer to denoise low-intensity adversarial examples, while \emph{Feature Squeezing} \cite{xu2017feature} combines the method of \emph{adversarial training}. Although they perform well in experiments, they have complicated structures or cumbersome mechanisms due to the addition of external defense methods to achieve hybrid defense, which makes them prohibitively impractical. In contrast, DAFAR does not require any external structures or methods to form hybrid defense. That is why DAFAR is an \emph{autogenous hybrid defense} method, which guarantees the simple structure and effortless mechanism of DAFAR.

\section{Preliminaries}\label{motivation}
\subsection{Motivation and Goal}
Despite the considerable research effort expended towards defending against adversarial examples, scientific literature still lacks universal and effective methods to defend against adversarial examples.

To achieve ideal effectiveness for adversary defense, Defense-GAN, MagNet, and Feature Squeezing provide preliminary examples. We advocate four principles to achieve ideal defense effectiveness.

\begin{enumerate}
\item
Instead of assuming knowledge of the specific process for generating the adversarial examples, find intrinsic common properties among all adversarial examples across attack methods and parameters, such as pixel perturbation, label changing, and feature space interference. This principle will lead to attack-independent detection and defense.\label{prin}
\item
For detection methods, first amplify the difference between normal sample and adversarial example as much as possible, use comparable features to express their difference, and detect adversarial examples according to that difference. Do not operate on raw pixel values. This principle will lead to high detection accuracy, especially on high-intensity attack.\label{prin1}
\item
For purification methods, destroy the structure of adversarial examples by finding a close legitimate sample to the adversarial example in manifold, or transforming the adversarial image. Specially crafted adversarial perturbations are usually not robust, especially those generated from weak attacks. Even tiny changes can destroy the attack effect. This principle will lead to effective defense on low-intensity and low-robust attack.\label{prin2}
\item
Use a hybrid of detection and purification to achieve the ideal defense effect within the full range of attack intensity.\label{prin3}
\end{enumerate}

Our goal is to design an adversary defense framework based on the above four principles, which can achieve high effectiveness as well as universality across attack methods and parameters, with relatively low area and time overhead. To do so, we propose DAFAR to defend deep learning models against adversarial examples in an autogenous-hybrid-defense manner, based on common properties of adversarial examples, i.e., feature space interference. We carry out empirical experiments to demonstrate the rationale of our ideas. We also conduct experiments to evaluate DAFAR by comparing it with several representative and state-of-the-art adversary defense methods.

\subsection{Threat Model}
As done in many prior works \cite{carlini2017adversarial, athalye2018obfuscated, biggio2013evasion, 8482346}, we give three threat models in this paper:

\begin{enumerate}
    \item A \emph{Black-box Adversary} generates adversarial examples without the knowledge of the victim model (i.e., the training methods, the model architecture and parameters) and the defense method.
    
    \item A \emph{Gray-box Adversary} generates adversarial examples with the knowledge of unsecured victim model but is not aware that the defense method is in place.
    
    \item A \emph{White-box Adversary} generates adversarial examples with the full knowledge of victim model and the defense method (i.e., the training methods, the model architecture and parameters, the defense structure, parameters and mechanism). It is also often called adaptive attack or defense-aware attack.
\end{enumerate}

We mainly consider the \emph{Gray-box Adversary} in our design and experiments, because in actual software application scenarios, due to confidentiality and product packaging, it is difficult for attackers to get all the information about the victim model and its defenses. And because of the independent plug-in property of the feedback network and the detector, they can be easily disassembled or installed from the victim model without affecting its performance, so they are not regarded as a part of the threat model. 

\subsection{Datasets}
In this paper we consider two datasets used throughout the existing work in this field.

The \emph{MNIST} dataset \cite{lecun1998mnist} consists of $70,000$ $28\times28$ greyscale images of handwritten digits from $0$ to $9$. Our CNN achieves $99.14\%$ accuracy on this dataset.

The \emph{CIFAR-10} dataset \cite{krizhevsky2009learning} consists of $60,000$ $32\times32$ color images of ten different objects (e.g., horse, airplane, etc). This dataset is substantially more difficult. For representativeness, we use a common DNN architecture as the victim model. It achieves an $86.17\%$ accuracy, which is at the normal level.

\section{Principle and Design}\label{secdesign}
\subsection{High-level Feature Interference}\label{feature}

\begin{figure}[h]
\begin{center}
  \includegraphics[width=1\linewidth]{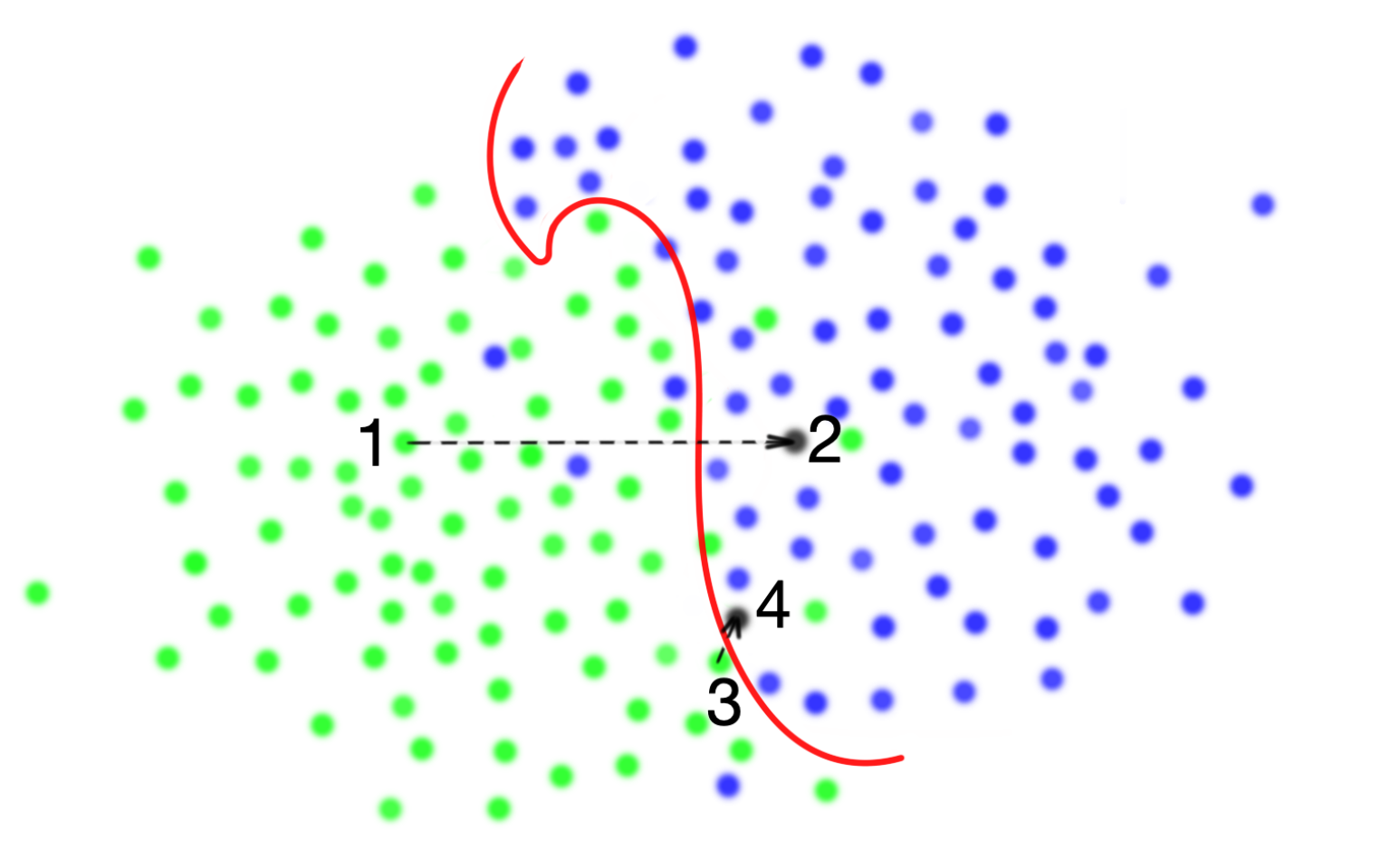}
\end{center}
   \caption{2-D sample space. The red curve is the boundary of the manifold of the task. The green points and blue points are separately in two categories, while the gray points are adversarial examples.}
\label{fig:manifold}
\end{figure}

Whether gradient-based or optimization-based attacks misguide the classifier by adding specially crafted small perturbations to clean samples. Deep learning models determine the label of a sample by extract the high-level features of the sample at deep layers \cite{LIU201711}. There are two cases how adversarial examples mislead a classifier \cite{meng2017magnet}:
\begin{enumerate}
\item
The original sample is far from the boundary of the manifold of the task, shown as the green point 1 in Figure \ref{fig:manifold}. Adversarial perturbations significantly interfere in the feature extraction process, namely, strong attack, inducing huge disturbance into high-level features extracted, leading to unexpected change of feature semantics, ultimately causing the classifier to mis-classify, shown as the gray point 2 in Figure \ref{fig:manifold}.
\item
The original sample is close to the boundary of the manifold, shown as the green point 3 in Figure \ref{fig:manifold}. Though adversarial perturbations just slightly interfere in the feature space, namely, weak attack, the classification easily changes because of its original low confidence, shown as the gray point 4 in Figure \ref{fig:manifold}.

\end{enumerate}

In the first case, adversarial perturbations significantly interfere in the feature space. Formally, given $f(\cdot)$ as the victim classifier, $E(\cdot)$ as feature extraction layers of $f$, $F(\cdot)$ as the output layers of $f$, $x$ as a normal sample, and $x'$ as an adversarial sample of $x$, the process that $x'$ misguides $f$ can be described as
\begin{equation}\label{semantics}
\begin{aligned}
\delta(E(x), E(x'))&\gg \delta(x,x')\\
F(E(x'))&\neq F(E(x))
\end{aligned}
\end{equation}
where $\delta(a,b)$ means the difference between $a$ and $b$ for a given distance function $\delta(\cdot,\cdot)$. We will demonstrate the high-level feature interference caused by adversary in Section \ref{secinterference} with experiments.

In the second case, adversarial perturbations just slightly interfere in the feature space, so the process that $x'$ misguides $f$ can be described as
\begin{equation}\label{semantics2}
\begin{aligned}
\delta(E(x), E(x'))&\approx 0\\
F(E(x'))&\neq F(E(x))
\end{aligned}
\end{equation}

\subsection{Reconstruction Errors}\label{secerror}
In Formula \ref{semantics}, when encountering relatively strong attacks, we find an inequality, $\delta(E(x), E(x'))\gg \delta(x,x')$, corresponding to \emph{Principle \ref{prin1}} that to detect adversarial examples with high accuracy, the amplification of the difference between normal sample and adversarial example is necessary.

What we have when detecting are an input sample $x$, deep neural network $f$ and high-level features $E(x)$. According to \cite{hinton2007learning, 10.1145/1390156.1390294, BELLO2020259}, an autoencoder reconstructs a sample using the features extracted by its encoder (feature extraction layers). If the features extracted by encoder are disturbed, the decoded output will present a significant \emph{reconstruction error}. This significant reconstruction error, or called \emph{reconstruction distance} between the reconstructed sample and the original sample is exactly what we want, according to \emph{Principle \ref{prin1}}. So we can add a feedback decoder $D(\cdot)$ to the feature extraction layers $E(\cdot)$ of victim network, to reconstruct the high-level features $E(x)$ to a reconstruction $D(E(x))$. The victim network $E(\cdot)$ and the feedback network $D(\cdot)$ constitute a feedback autoencoder $D(E(\cdot))$. This structure transforms the attack on the victim network (imperceptible perturbations) into an obvious attack on the feedback autoencoder (reconstruction error) directly, because the feedback autoencoder and the victim network share the interference introduced by possible adversarial perturbations in high-level feature space as they share the feature extraction layers. After appropriate training, the normal samples will be reconstructed perfectly with small reconstruction errors, while the adversarial ones that attack the victim network will present significant reconstruction errors because they \emph{attack the feedback autoencoder as well}. Formally we can give a description by
\begin{equation}
\begin{aligned}
\delta(x',D\left(E(x')\right))&\gg \delta(x, D\left(E(x)\right))
\end{aligned}
\end{equation}
where $\delta(x, D\left(E(x)\right))$ is the reconstruction distance of sample $x$, which can be described as
\begin{equation}
\delta(x, D\left(E(x)\right)=||x-D\left(E(x)\right)||_p
\end{equation}

In this way we greatly amplify the difference between a normal sample and an adversarial example. By detecting the difference, we can detect adversarial examples with higher accuracy. Moreover, since the amplification is based on the common properties among all adversarial examples, according to \emph{Principle \ref{prin}}, this method is attack-independent. In other words, DAFAR is able to detect adversarial examples with high accuracy and universality. More formally, we describe detection process of DAFAR by
\begin{equation}
O(x) =F(E(x))\land C\left(\delta(x, D\left(E(x)\right))\right)
\end{equation}
where $C(\cdot)$ is the detector that judges a sample whether legitimate or adversarial by the reconstruction errors, which can optionally be an anomaly detector.

Compared to the anomaly detector of MagNet, DAFAR detects the disturbance caused by adversarial examples in the victim network fully, for it directly changes the attack on victim network to the attack on the feedback autoencoder, through shared feature extraction layers.

\subsection{Purification by Reconstruction}\label{sec:PurifybyRecon}
In Formula \ref{semantics2}, when encountering very weak attacks, the high-level feature disturbance is so small that the feedback process cannot generate large enough reconstruction errors to be detected. However, the feedback autoencoder is also a generative model, so it has a high probability to find a close legitimate sample to the weak adversarial example in manifold, just as Defense-GAN \cite{samangouei2018defense}. Formally, the generative defense process can be described by
\begin{equation}
\begin{aligned}
\delta(x, D(E(x')))&\approx 0\\
F(E(D(E(x'))))&\approx F(E(x))
\end{aligned}
\end{equation}

In aspect of adversarial examples' robustness to explain it, because adversarial perturbations are crafted by machine learning methods, most adversarial examples, especially weak ones, have poor robustness. Although the reconstruction process tries to make the reconstructed sample close to the original sample, there will always be changes in the pixel value, which will destroy the structure of the adversarial perturbations and make the adversarial perturbations invalid.

\subsection{DAFAR Structure}
\begin{figure}[t]
\begin{center}
  \includegraphics[width=1\linewidth, trim=130 70 240 120,clip]{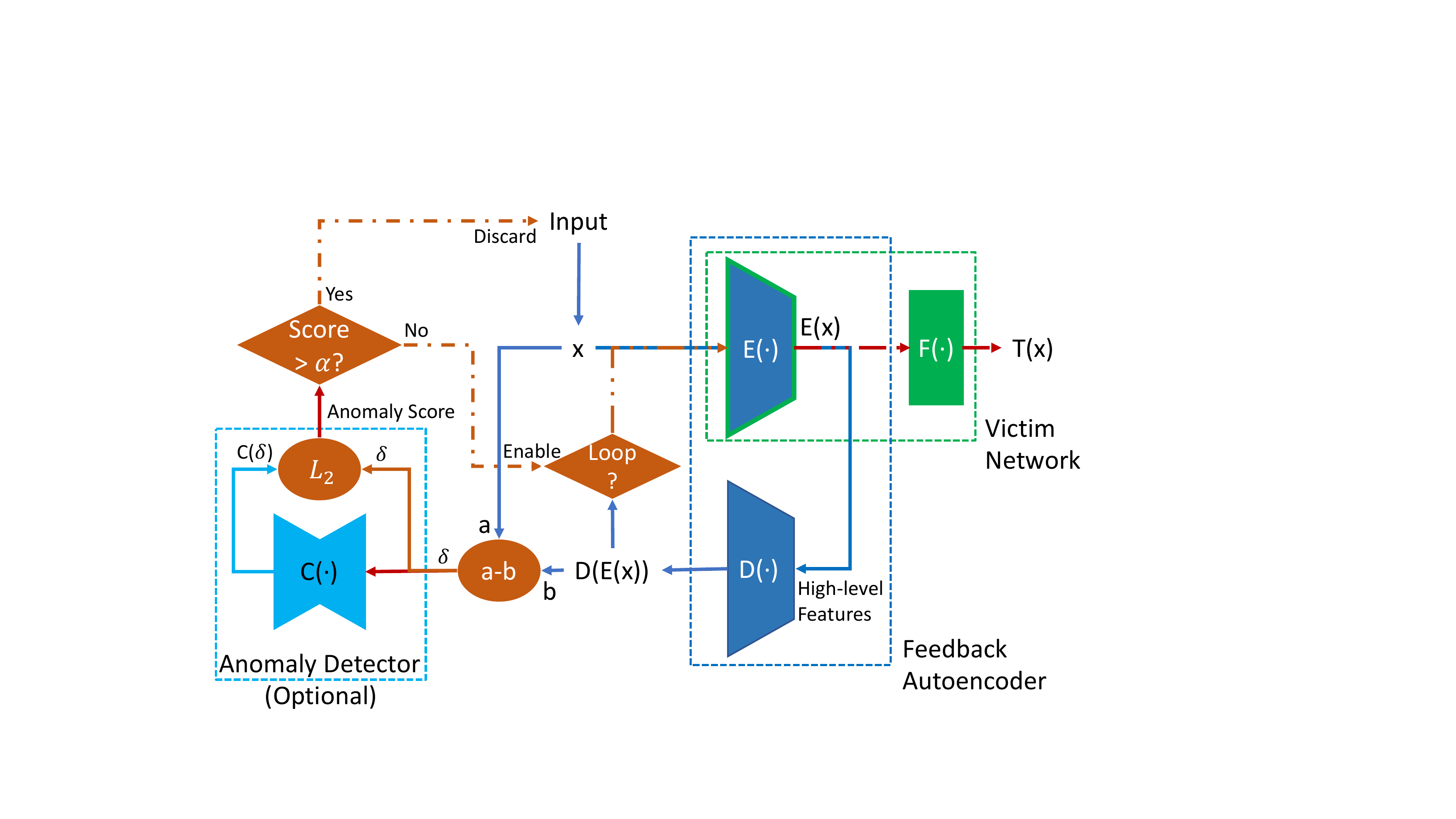}
\end{center}
   \caption{Detailed structure and workflow of DAFAR. Feedback network $D(\cdot)$ reconstructs the input $x$ from high-level features $E(x)$, and then the detector $C(\cdot)$ gives an anomaly score in the form of $L_2$ distance for the input $x$, according to the reconstruction errors $\delta$, and judges whether the input is adversarial by comparing the score with a threshold $\alpha$. The reconstruction of dis-adversarial judgments will be put back into the victim network for classification, and adversarial judgements will be discarded.}
\label{fig:structure}
\end{figure}

We discuss the detailed structure and workflow of DAFAR in this section, which is shown in Figure \ref{fig:structure}.

\subsubsection{Victim Network} 
A victim network is the deep learning model facing adversarial attacks directly, shown as $F(E(\cdot))$ in Figure \ref{fig:structure}. The encoder $E(\cdot)$ of feedback autoencoder $D(E(\cdot))$ is exactly the feature extraction layers $E(\cdot)$ of the victim network $F(E(\cdot))$. This part of DAFAR transforms the disturbance in the feature extraction layers of victim network directly into disturbance in the encoder of feedback autoencoder, which will cause significant reconstruction errors in the decoder $D(\cdot)$ later when encountering relatively strong attacks.

\subsubsection{Decoder/Feedback Network}
The feedback network, which is also called decoder and shown as $D(\cdot)$ in Figure \ref{fig:structure}, reconstructs the input $x$ from high-level features extracted by the encoder ($E(x)$). Though the victim network and the decoder are two separated parts, to ensure the reconstruction quality and the classifying accuracy of victim classifier, we train victim network and feedback network at the same time by minimizing a loss function over the training set, where the loss function is the combination of cross entropy function loss and mean squared error loss
\begin{equation}
\begin{aligned}
J_f(&\mathbb{X}_{train})=\frac{1}{\# \mathbb{X}_{train}}\cdot \sum_{x\in\mathbb{X}_{train}} \\
&\left(||x-D(E(x))||_2+\sum p(x)log\frac{1}{F(E(x))}\right)
\end{aligned}
\end{equation}
where $\mathbb{X}_{train}$ is the training dataset only containing normal samples, and $p(x)$ is the grand truth label vector of sample $x$. Training in this manner will not affect the accuracy of original victim network, for autoencoder is often used in pre training and shows effectiveness in learning representations for subsequent classification tasks \cite{kokalj2019mitigation, 10.1007/978-981-15-6353-9_34}, which we will demonstrate in our experiments later.

\subsubsection{Detector}\label{detector} 
As we mentioned above, DAFAR distinguishes adversarial examples by detecting significant reconstruction distance. We use an anomaly detection autoencoder as the decoder, shown as $C(\cdot)$ in Figure \ref{fig:structure}. Autoencoder is often used in anomaly detection \cite{10.1007/978-3-030-50516-5_13, 9248108, 9113298}. In most cases, the number of adversarial examples is much less than normal samples, so adversarial examples are an \emph{anomaly}. However, since the difference between normal samples and adversarial examples is very small, directly applying anomaly detection autoencoder to adversary detection is not recommended. Yet in our case, we have greatly amplified the difference between adversarial examples and normal samples by feedback autoencoder, in the form of reconstruction errors. So it is reasonable to use anomaly detection autoencoder as our final detector. To this end, we use \emph{image subtraction} ($a - b$ in Figure \ref{fig:structure}) as reconstruction errors. Given $\mathbb{E}_{train}$ as the the training set only containing reconstruction errors of normal samples, and $\delta$ as the reconstruction errors of a normal sample in $\mathbb{E}_{train}$, we train the detector $C(\cdot)$ by minimizing mean squared error loss in a semi supervised learning manner
\begin{equation}
J_C\left(\mathbb{E}_{train}\right)=\frac{1}{\# \mathbb{E}_{train}}\cdot\sum_{\delta\in\mathbb{E}_{train}}||\delta-C(\delta)||_2
\end{equation}

After training, the detector $C(\cdot)$ will reconstruct normal data well, while failing to do so for anomaly data which the detector $C(\cdot)$ has not encountered. The detector $C(\cdot)$ does not tell us whether a reconstruction error belongs to an adversarial example or not directly. It only gives us a reconstruction distance between its input and output, which is used as \textbf{\emph{anomaly score}}. Input with high anomaly score is considered to be an anomaly. In order to define whether an anomaly score is high or low, a threshold $\alpha$ is needed. We set the $99.7\%$ confidence interval's right edge calculated from anomaly scores of all normal samples in training set as $\alpha$ for MNIST, and $95\%$ for CIFAR-10. Given $\bar{x}$ as the average score, $\sigma$ as the standard deviation and $n$ as the number of normal samples in training set, then $\alpha$ is
\begin{equation}
\alpha = \bar{x}+z\cdot\frac{\sigma}{n}, \quad z = 2, 3
\end{equation}
For other datasets, people can set the threshold with an appropriate confidence interval according to real needs, in the same manner as MNIST and CIFAR-10 here.

However, note that the anomaly detector here is optional in order to reduce area and time overhead when actually used in software. Because instead of using the anomaly detector, we can directly calculate the $L_2$ distance of reconstruction errors and use the same threshold as mentioned above to detect adversaries. Although the detection accuracy may be reduced, because DAFAR is a autogenous hybrid defense method and there is the feedback purification afterwards, discarding the detector has only small influence on the final defense effect, which we will show in our experiments later. But if pursuing the most effective defense, we recommend adding this anomaly detector.

\section{Experiments}\label{secexp}
\subsection{Network Structure}
In this section we describe network structures used in our experiments.

\textbf{Victim network.} We choose common network structures as victim networks, which face the adversarial attacks directly, shown in Table 1. Though we have mentioned the encoder of feedback autoencoder is the feature extraction layers of victim network, actually the encoder does not have to include all feature extraction layers. It can be just several former layers, according to real needs. In other words, we can determine the layers to capture high-level features (i.e., feedback positions) according to what trade-off we want to make between training overhead and detection effectiveness.
 \begin{table}[h]
\begin{center}
\scalebox{1}{\begin{tabular}{|ll|ll|}
\hline
MNIST $\mathcal{T}_M$ && CIFAR-10 $\mathcal{T}_C$& \\
\hline\hline
Encoder $\mathcal{E}_M$ &&Encoder $\mathcal{E}_C$& \\
\hline
Conv.ReLU & $3\times 3 \times 32$ &Conv.ReLU  & $3\times 3 \times 96$\\
Conv.ReLU & $3\times 3 \times 32$ &Conv.ReLU  & $3\times 3 \times 96$\\
MaxPool & $2\times 2$  &Conv.ReLU  & $3\times 3 \times 96$ \\
Conv.ReLU & $3\times 3 \times 64$ &MaxPool & $2\times 2$\\
Conv.ReLU & $3\times 3 \times 64$ &Conv.ReLU  & $3\times 3 \times 192$\\
MaxPool & $2\times 2$&Conv.ReLU  & $3\times 3 \times 192$ \\
&  &Conv.ReLU  & $3\times 3 \times 192$ \\
&  & MaxPool & $2\times 2$ \\
\hline
Output $\mathcal{F}_M$ &&Output $\mathcal{F}_C$& \\
\hline
Linear.ReLU & $200$ &Conv.ReLU  & $3\times 3 \times 192$\\
Linear.ReLU & $200$ &Conv.ReLU  & $1\times 1 \times 192$\\
Softmax & $10$ &Conv.ReLU  & $1\times 1 \times 10$\\
&  &Linear.ReLU  &$200$\\
&  &Linear.ReLU  &$200$\\
&  &Softmax & $10$  \\
\hline
\end{tabular}}
\end{center}
\caption{Structures of victim networks.}
\end{table}

\textbf{Decoder/Feedback network.} The structures of decoders are often the reverse of their encoders, shown in Table 2. But they can be modified according to real needs, such as transforming from the byproduct network in unsupervised pre training.

 \begin{table}[h]
\begin{center}
\scalebox{0.9}{\begin{tabular}{|ll|ll|}
\hline
MNIST $\mathcal{D}_M$ && CIFAR-10 $\mathcal{D}_C$& \\
\hline\hline
MaxUnpool & $2\times 2$ & MaxUnpool & $2\times 2$\\
ConvTranspose.ReLU & $3\times 3 \times 64$ &ConvTranspose.ReLU  & $3\times 3 \times 192$\\
ConvTranspose.ReLU & $3\times 3 \times 32$ &ConvTranspose.ReLU  & $3\times 3 \times 192$\\
MaxUnpool & $2\times 2$  &ConvTranspose.ReLU  & $3\times 3 \times 96$ \\
ConvTranspose.ReLU & $3\times 3 \times 32$ &MaxUnpool & $2\times 2$\\
ConvTranspose.Tanh & $3\times 3 \times 1$ &ConvTranspose.ReLU  & $3\times 3 \times 96$\\
&  &ConvTranspose.ReLU  & $3\times 3 \times 96$ \\
&  &ConvTranspose.Tanh  & $3\times 3 \times 3$ \\
\hline
\end{tabular}}
\end{center}
\caption{Structures of decoders.}
\end{table}

\textbf{Detector.} Detectors are simple 5-layer fully connected autoencoders for MNIST and CIFAR-10, shown in Table 3. Note that we did not carefully craft the structure of the detectors, for the input patterns are very simple. But the experimental results later are outstanding, showing the effectiveness of DAFAR. 

 \begin{table}[h]
\begin{center}
\scalebox{1}{\begin{tabular}{|ll|ll|}
\hline
MNIST $\mathcal{C}_M$&& CIFAR-10 $\mathcal{C}_C$& \\
\hline\hline
Linear.ReLU & $256$ &Linear.ReLU  & $512$\\
Linear.ReLU & $32$ &Linear.ReLU & $64$\\
Linear.ReLU & $256$ &Linear.ReLU  & $512$\\
Linear.Tanh & $784$ &Linear.Tanh  & $3072$\\
\hline
\end{tabular}}
\end{center}
\caption{Structures of detectors}
\end{table} 

\subsection{Reconstruction Error: How Does DAFAR Detect?}\label{secwhy}
In this section we conduct experiments to characterize the phenomenon of reconstruction errors in DAFAR structure, to explain how DAFAR detects.

\subsubsection{High-level Feature Interference}\label{secinterference}
As we discussed in Section \ref{feature}, whether gradient-based or optimization-based attacks induce huge disturbance into high-level features, leading to unexpected changes of feature semantics. In order to show high-level feature interference caused by adversarial perturbations, we carry out experiments by extracting high-level features of adversarial examples and their original normal samples from deep layers of victim network and calculating distance between them. We compare the result with the distance between high-level features of samples added same-intensity Gaussian noise and that of their original samples in Table 4. 

\begin{table}[h]
\begin{center}
\label{tab:highlevel}
\scalebox{1}{\begin{tabular}{|c|c|c|c|}
\hline
Perturbations &Gaussian $(0, 0.3)$ & FGSM $0.3$& PGD $0.3$ \\
\hline\hline
$L_2$ distance&$36.13$ & $102.64$ & $96.44$ \\
\hline
\end{tabular}}
\end{center}
\caption{Average high-level features $L_2$ distance between samples with certain perturbations and their original samples, of $1000$ samples in MNIST test set.}
\end{table}

Clearly adversarial perturbations cause a much bigger interference to high-level features of a sample. Here we can draw an observation.

\textbf{Observation 1.} \emph{Adversarial perturbations induce huge disturbance into high-level features extracted by deep layers of victim network.}

\begin{figure}[b]
\begin{center}
  \includegraphics[width=1\linewidth]{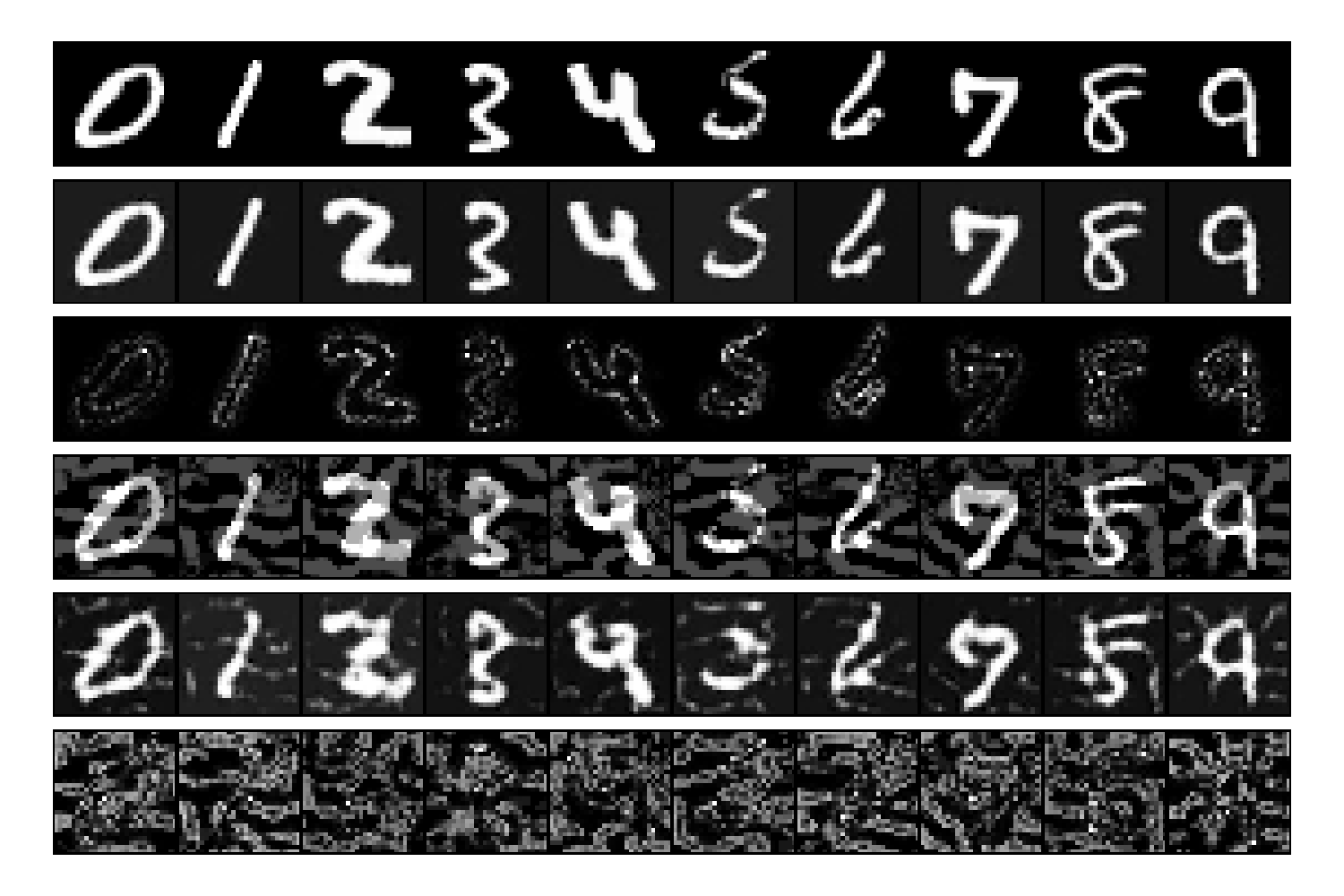}
\end{center}
   \caption{Reconstruction errors of MNIST normal and adversarial samples. The first two lines are separately normal samples and their reconstructions, and the third line are their reconstruction errors. The forth and fifth lines are separately adversarial examples and their reconstructions, and the last line are their reconstruction errors. There is significant difference between the third and last lines, in the aspects of errors size and patterns.}
\label{fig:recoerrors}
\end{figure}

\subsubsection{Reconstruction Errors}
The interference in adversarial examples' high-level features will lead to big reconstruction errors, as we discussed in Section \ref{secerror}. In this section we will give more details on characterizations of reconstruction errors.

\begin{figure*}[h]
\begin{center}
  \includegraphics[width=1\linewidth]{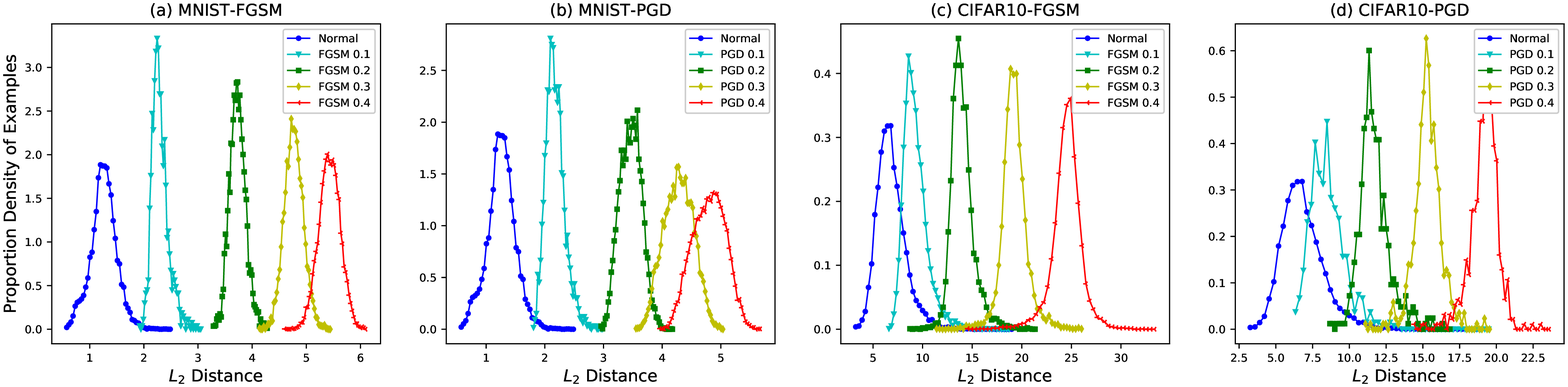}
\end{center}
   \caption{Reconstruction-distance proportion density curves of adversarial examples and normal samples, across different attack methods and intensities. Every peak shows the area of $L_2$ distance concentration of a certain attack intensity. The separated peaks show reconstruction distance increases with attack intensity, which also causes adversarial examples differentiated from normal samples. However, in CIFAR-10 the distinction between the normal samples' peak and low-intensity adversarial examples' peak is not that clear.}
\label{fig:recoerrors2}
\label{fig:onecol}
\end{figure*}

Figure \ref{fig:recoerrors} shows the significant difference between the reconstruction errors of normal samples and that of adversarial examples clearly. To quantitatively characterize how reconstruction errors distribute with attack methods and intensities, we calculate the reconstruction distance of normal samples and adversarial examples across different attack methods and intensities in the form of $L_2$ distance. The results are shown in Figure \ref{fig:recoerrors2}. 

Here we can draw three observations.

\textbf{Observation 2.} \emph{Difference of reconstruction errors between normal and adversarial samples is significant in two aspects: 1) reconstruction errors of normal and adversarial samples show very different patterns; 2) reconstruction errors in the form of $L_2$ shows distinctively separated concentrations across attack intensities.}

\textbf{Observation 3.} \emph{Though the difference is clear, there is no fixing reconstruction-error threshold to perfectly divide normal samples and adversarial examples, especially at low attack intensity.}

\textbf{Observation 4.} \emph{Difference of reconstruction errors between adversarial examples and normal samples increases with attack intensities.}

\subsubsection{Anomaly Score}\label{SecAnomalyScore}
\begin{figure*}[htb]
\begin{center}
  \includegraphics[width=1\linewidth]{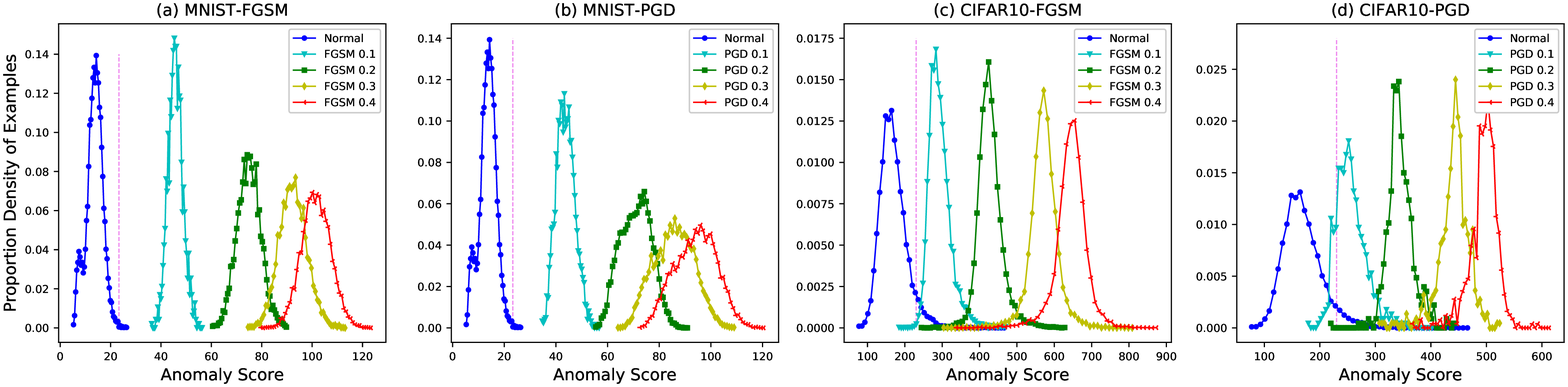}
\end{center}
   \caption{Anomaly-score proportion density curves of adversarial examples and normal samples, across different attack methods and intensities. Every peak shows the area of anomaly score concentration of a certain attack intensity. Violet vertical line indicates the location of the anomaly score threshold, which clearly divides normal samples and adversarial examples.}
\label{fig:anomaly}
\end{figure*}

As we discussed in Section \ref{detector}, detector needs a threshold to tell whether an input is an anomaly, so there should be a clear dividing line of anomaly score between normal inputs and anomalies. We train a detector in semi supervised manner on clean samples' reconstruction errors. Then we input clean samples and adversarial examples across different attack methods and intensities to calculate their anomaly scores. Figure \ref{fig:anomaly} shows how anomaly scores distribute with attack methods and intensities. Here we draw two important observations.

\textbf{Observation 5.} \emph{The distribution of anomaly scores is approximately normal, so we assume that the score follows a normal distribution. We set the $99.7\%$ confidence interval's right edge of normal samples' anomaly scores for MNIST and $95\%$ for CIFAR-10 as the threshold to distinguish normal and adversarial samples, as discussed in Section \ref{detector}. We show the score threshold of MNIST and CIFAR-10 in Table 5, and also show in Figure \ref{fig:anomaly}.}
 \begin{table}[h]
\begin{center}
\label{tab:threshold}
\scalebox{1}{\begin{tabular}{|c|c|}
\hline
Dataset & Score threshold\\
\hline\hline
MNIST & $23.333$\\
\hline
CIFAR-10 & $230.143$\\
\hline
\end{tabular}}
\end{center}
\caption{Score threshold of MNIST and CIFAR-10.}
\end{table}

\textbf{Observation 6.} \emph{Detector further magnifies the difference between normal and adversarial samples. There is a clear dividing line of anomaly score between normal input and anomaly, which means it is reasonable to determine a threshold to tell whether an input is an anomaly.}

We hypothesize that detector's secondary amplification effect is introduced by two reasons: 1) reconstruction distances of adversarial examples with large intensities are much bigger than that of normal samples, which detector can easily distinguish; 2) even if the reconstruction-distance difference between normal and adversarial samples is not that much, their reconstruction errors show very different patterns as shown in Figure \ref{fig:recoerrors}, which detector can refer to.

\subsection{Evaluation of Detection}\label{seceval}
In this section we evaluate the accuracy and universality of DAFAR in detecting adversarial examples using FGSM, JSMA, $\rm CW_2$ and PGD across different attack intensities, and compare the results with only-binary-classifier method, detection system of MagNet and Feature Squeezing. For FGSM, PGD and $\rm CW_2$, we used the implementation of Cleverhans \cite{2016cleverhans}. For JSMA, we use authors’ open source implementation \cite{papernot2016limitations}. 

In principle, DAFAR shows a better performance of accuracy than MagNet, and Feature Squeezing, especially in low attack intensities, and the same level of universality as MagNet and Feature Squeezing across attack methods, which is much better than only-binary-classifier method.

\subsubsection{Detection Accuracy and Universality across Intensities}
In this section we evaluate DAFAR's adversary-detection accuracy and universality across different attack intensities, separately on MNIST and CIFAR-10, using FGSM attack across different attack intensities.\\

\begin{figure}[htb]
\begin{center}
\includegraphics[width=1\linewidth]{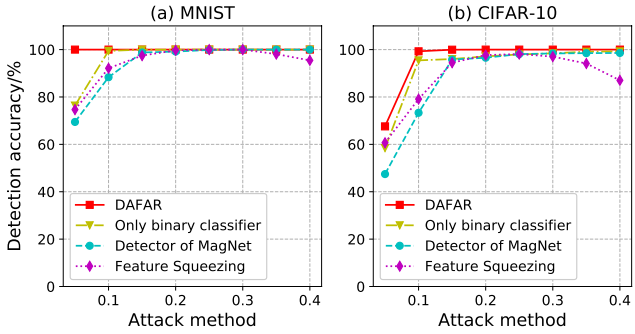}
\end{center}
   \caption{Detection accuracy curves on MNIST and CIFAR-10 adversarial examples across different FGSM intensities.}
\label{fig:accintens}
\end{figure}

\textbf{MNIST.} We train a victim network in DAFAR method on MNIST and achieve an accuracy of $99.24\%$ on the test set, which is close to the state of the art. We test the adversary detection accuracy of each method on test sets only containing FGSM adversarial examples across different attack intensities. The results are shown in Figure \ref{fig:accintens}. Here we can draw some conclusions.

\emph{Effect on normal examples.} The victim network trained in DAFAR method achieves an accuracy of $99.24\%$, and the detector of DAFAR shows a false positive rate of only $0.16\%$, which means DAFAR does not affect victim network's accuracy.

\emph{Effect on adversarial examples.} DAFAR detects MNIST adversarial examples in an accuracy of $100\%$ across all attack intensities, as shown in Figure \ref{fig:accintens}, higher than other three methods especially in low attack intensities, showing the best detection accuracy and universality. Because DAFAR can eliminate all MNIST adversarial examples by detection alone, the subsequent evaluations of DAFAR hybrid defense in Section \ref{eva:hybrid} are only conducted on CIFAR-10.\\

\textbf{CIFAR-10.} CIFAR-10 is a much more complex dataset than MNIST. We train a victim network in DAFAR method on CIFAR-10 and achieve an accuracy of $86.17\%$ on the test set, which is at the normal level. We test the adversary detection accuracy of each method on test sets only containing FGSM adversarial examples across different attack intensities. The results are shown in Figure \ref{fig:accintens}. Here we can draw some conclusions.

\emph{Effect on normal examples.} The victim network trained in DAFAR method on CIFAR-10 achieves an accuracy of $86.17\%$, and the detector of DAFAR shows a false positive rate of only $3.49\%$ on normal samples. It is a negligible performance reduction.

\emph{Effect on adversarial examples.} DAFAR detects CIFAR-10 adversarial examples in an accuracy of $100\%$ across most of attack intensities, as shown in Figure \ref{fig:accintens}, but not as accurate as on MNIST when the attack intensity is very low. However, it is still higher than other three methods especially in low attack intensities. This also provides empirical evidence that DAFAR achieves the best detection effectiveness across different attack intensities.

\subsubsection{Detection Accuracy and Universality across Methods}
In this section we evaluate DAFAR's detection accuracy and universality across different attack methods, separately on MNIST and CIFAR-10, using FGSM, JSMA, $\rm CW_2$ and PGD.

\begin{figure}[h]
\begin{center}
\includegraphics[width=1\linewidth]{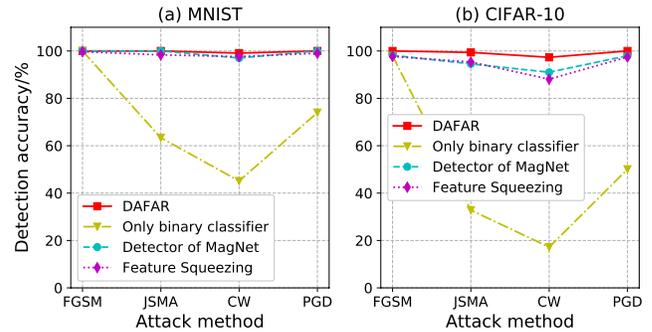}
\end{center}
   \caption{Detection accuracy curves on MNIST and CIFAR-10 adversarial examples across different attack methods.}
\label{fig:accmed}
\end{figure}

\textbf{MNIST.} We test the adversary detection accuracy of each method on MNIST test sets containing adversarial examples across different attack methods. Figure \ref{fig:accmed} shows the results, which provides evidence that DAFAR has the same level of universality across different attack methods as detection system of MagNet and Feature Squeezing, much better than only-binary-classifier method.

\textbf{CIFAR-10.} We test the adversary detection accuracy of each method on CIFAR-10 test sets containing adversarial examples across different attack methods. The results are shown in Figure \ref{fig:accmed}, which give the same conclusion as MNIST's.

For all parts in DAFAR are trained in semi supervised way or only on clean samples, theoretically DAFAR has outstanding universality across attack methods, which is strongly proved by our experimental results. Note that DAFAR detection alone can achieve that considerable defense effect.

\subsection{Evaluation of Purification}\label{eva:purify}
\begin{figure}[ht]
\centering
\includegraphics[width=1\linewidth]{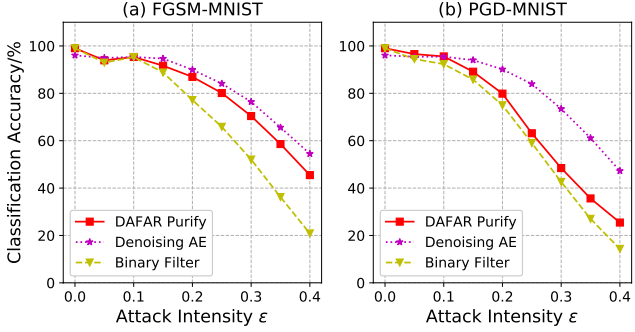}
\caption{Classification accuracy curves on MNIST adversarial data-sets across attack intensities.}
\label{fig:purify}
\end{figure}

In Section \ref{sec:PurifybyRecon} we discuss the mechanisms how feedback reconstruction purifies weak adversarial examples in aspects of generative model and adversarial examples' robustness. In this section we conduct experiments to evaluate the defense effect of DAFAR purification.

As shown in Figure \ref{fig:purify}, we compare DAFAR purification with the generative method --- denoising autoencoder, and the input transformation method --- binary filter. Though DAFAR's feedback autoencoder is trained by only normal samples, it shows a purification effect close to that of the denoising autoencoder on the FGSM adversarial examples. And on PGD adversarial examples, the purification effect of DAFAR shows good effectiveness when the attack intensity is low, which will make sense in DAFAR's autogenous hybrid defense. Furthermore, when the attack intensity is $0$, namely, the input normal samples, although there is one more feedback loop than the original classification data stream, the classification accuracy is $99.04\%$, very close to the original $99.14\%$. Therefore, the additional feedback reconstruction loop of DAFAR purification will not affect the performance of the victim model on legitimate samples.

\subsection{Evaluation of Autogenous Hybrid Defense}\label{eva:hybrid}

Though DAFAR detection shows a relatively good performance, it still does not achieve ideal detection accuracy at very low attack intensities when encountering complex data-sets. There are two ways to approach ideal defense effectiveness. First, train DAFAR in more appropriate parameters, network architectures and training methods, to compress anomaly score interval of normal samples (i.e., the blue peaks in Figure \ref{fig:anomaly}) as much as possible until the interval converges to $0$, which is the ideal condition, but very difficult to achieve. Second, find a way to deal specifically with adversarial examples with low attack intensity.

\begin{figure}[h]
\begin{center}
  \includegraphics[width=1\linewidth]{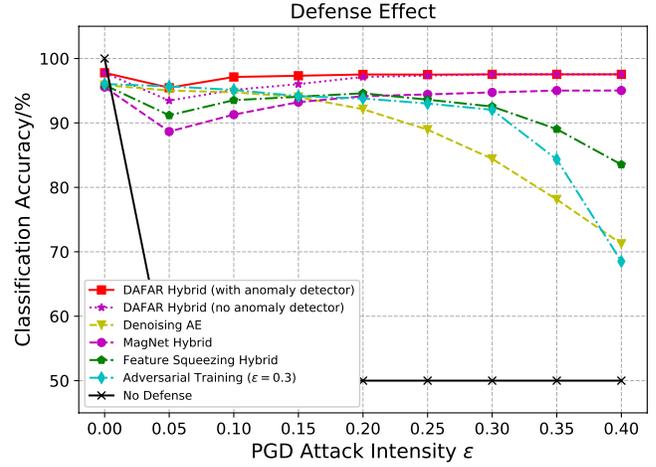}
\end{center}
   \caption{Classification accuracy curves on CIFAR-10 adversarial and normal samples $(1:1)$.}
\label{fig:DAFARs}
\end{figure}

Fortunately, according to \emph{Principle \ref{prin3}}, and as we discussed in Section \ref{sec:PurifybyRecon}, DAFAR's feedback autoencoder can be also used as a generative model, which is good at handling low-intensity adversaries. And as we obtained from experiments in Section \ref{eva:purify}, DAFAR purification indeed shows considerable effectiveness when attack intensity is low. So we compose DAFAR detection and purification to form an \emph{autogenous hybrid defense}. To evaluate this structure, we carry out experiments on data-sets derived from CIFAR-10 by conducting PGD attacks on CIFAR-10 samples that can be correctly classified by the victim model. In these data-sets, the ratio of adversarial examples and normal samples is $1:1$. We show the results in Figure \ref{fig:DAFARs}. The classification accuracies of DAFAR Hybrid across all attack intensities are near $100\%$, higher than all the other methods, presenting ideal defense effectiveness and low side effect. Even without the anomaly detector, DAFAR still shows considerable effectiveness.

\subsection{Evaluation of Practicality}\label{eva:efficiency}
\begin{table*}[ht]
\begin{center}
\label{tab:practicality}
\scalebox{1}{\begin{tabular}{|c|c|c|c|c|c|c|}
\hline
Method & DE & DU & $\#$ P & L & IO & Cost \\
\hline\hline
\tabincell{c}{DAFAR without\\ Anomaly Detector}  & High & High & $\frac{n}{2}$ & $n$ & None & Low \\ 
\hline
\tabincell{c}{DAFAR with\\ Anomaly Detector}  & High & High & $n$ & $\frac{3}{2}n$ & \tabincell{c}{Train the anomaly detector on\\ reconstruction errors} & Low \\ 
\hline
MagNet Hybrid & Middle & High & $> 2n$ & $2n$ & \tabincell{c}{Train autoencoders on\\ normal samples,\\choose $T$} & High \\
\hline
Feature Squeezing Hybrid & Middle & Middle & $0$ & $2n$ & \tabincell{c}{choose squeezing patterns,\\adversarial training} & High\\
\hline
Denoising Autoencoder & Middle & Middle & $n$ & $n$ & \tabincell{c}{Train an autoencoder} & Middle\\
\hline
Defense-GAN & Middle & High & $2n$ & $n$ & \tabincell{c}{Train a GAN} & High\\
\hline
Adversarial Training & Middle & Low & $0$ & $0$ & \tabincell{c}{Generate Adversarial Examples,\\Retraining} & High\\
\hline
Binary Classifier & High & Low & $n$ & $n$ & \tabincell{c}{Generate Adversarial Examples,\\Train a binary classifier} & Low\\
\hline
Binary Filter & Low & Middle & $0$ & $0$ & \tabincell{c}{Design the Filter} & Low\\
\hline
\end{tabular}}
\end{center}
\caption{Comparison of practicality between DAFAR and other representative defense methods.}
\end{table*}

To defend software mounted with deep learning system against adversarial examples, the defense method should be flexible, area-efficient, low-latency and low-cost, while guarenteeing a high defense effectiveness. However, almost all existing adversarial example defense methods fail to meet these requirements. They either have complicated structure or mechanism that cause prohibitively high overhead and latency, or the defense effect is not good enough, or the installation and implementation costs a lot, preventing them from being practically applied in actual software. Nevertheless, DAFAR is excellent in all three aspects. To evaluate the practicality of DAFAR, referring to previous experiments, we comprehensively compare the defense effect (DE), defense universality (DU), number of extra parameters\footnote{Given that the number of parameters of the victim network is $n$.} ($\#$ P. Namely, area overhead), extra dataflow length (L. Namely, latency), implementation operations (IO. Namely, flexibility), and cost of DAFAR with other representative defense methods in Table 6. So far, DAFAR is the most suitable adversarial example defense method for software, especially mobile device software, which have high demind for low memory footprint and low latency.


\section{Conclusion}
DAFAR is the first \emph{autogenous hybrid defense} method that helps deep learning models to defend against adversarial examples effectively and efficiently. Besides the victim network, DAFAR only contains a plug-in feedback network and an optional anomaly detector. DAFAR imports the high-level features from the victim model’s feature extraction layers into the feedback network to reconstruct the input. Namely, the feature extraction layers of the victim network and the feedback network constitute a \emph{feedback autoencoder}. For strong attacks, DAFAR transforms the imperceptible attack on the victim model into the obvious reconstruction-error attack on the feedback autoencoder directly, which is much easier to detect; for weak attacks, the reformation process destroys the structure of adversarial examples. Considering all parts in DAFAR are trained on normal samples or in semi supervised way, DAFAR is attack-independent. Our experiments explain DAFAR's work mechanisms and show DAFAR can defend against state-of-the-art attacks in high effectiveness and universality, with low overhead. We compare DAFAR with seven representative defense methods and derive the conclusion: so far, DAFAR is the most lightweight, effective and universal adversarial example defense method for the security and reliability of software mounted with deep learning systems.

\ifCLASSOPTIONcompsoc
  \section*{Acknowledgments}
\else
  \section*{Acknowledgment}
\fi

The authors would like to thank the experts from HUAWEI for their insightful comments and advice. The work is supported by 2020 SJTU-HUAWEI TECH Cybersecurity Innovation Lab, YBN2019105168-SOW06, National Natural Science Foundation of China under Grand 61972249.

\ifCLASSOPTIONcaptionsoff
  \newpage
\fi



%
\bibliography{egbib}

\begin{thebibliography}{10}

\bibitem{szegedy2013intriguing}
Christian Szegedy, Wojciech Zaremba, Ilya Sutskever, Joan Bruna, Dumitru Erhan,
  Ian Goodfellow, and Rob Fergus.
\newblock Intriguing properties of neural networks.
\newblock {\em Computer ence}, 2013.

\bibitem{goodfellow2014explaining}
Ian~J. Goodfellow, Jonathon Shlens, and Christian Szegedy.
\newblock Explaining and harnessing adversarial examples.
\newblock {\em Computer ence}, 2014.

\bibitem{papernot2016limitations}
Nicolas Papernot, Patrick McDaniel, Somesh Jha, Matt Fredrikson, Z~Berkay
  Celik, and Ananthram Swami.
\newblock The limitations of deep learning in adversarial settings.
\newblock In {\em 2016 IEEE European symposium on security and privacy
  (EuroS\&P)}, pages 372--387. IEEE, 2016.

\bibitem{carlini2017towards}
Nicholas Carlini and David Wagner.
\newblock Towards evaluating the robustness of neural networks.
\newblock In {\em 2017 IEEE symposium on security and privacy (sp)}, pages
  39--57. IEEE, 2017.

\bibitem{madry2017towards}
Aleksander Madry, Aleksandar Makelov, Ludwig Schmidt, Dimitris Tsipras, and
  Adrian Vladu.
\newblock Towards deep learning models resistant to adversarial attacks.
\newblock {\em arXiv preprint arXiv:1706.06083}, 2017.

\bibitem{zhang2020generating}
Huangzhao Zhang, Hao Zhou, Ning Miao, and Lei Li.
\newblock Generating fluent adversarial examples for natural languages.
\newblock {\em arXiv preprint arXiv:2007.06174}, 2020.

\bibitem{kurakin2016adversarial}
Alexey Kurakin, Ian Goodfellow, Samy Bengio, et~al.
\newblock Adversarial examples in the physical world, 2016.

\bibitem{athalye2018synthesizing}
Anish Athalye, Logan Engstrom, Andrew Ilyas, and Kevin Kwok.
\newblock Synthesizing robust adversarial examples.
\newblock In {\em International conference on machine learning}, pages
  284--293. PMLR, 2018.

\bibitem{papernot2017practical}
Nicolas Papernot, Patrick McDaniel, Ian Goodfellow, Somesh Jha, Z~Berkay Celik,
  and Ananthram Swami.
\newblock Practical black-box attacks against machine learning.
\newblock In {\em Proceedings of the 2017 ACM on Asia conference on computer
  and communications security}, pages 506--519, 2017.

\bibitem{meng2017magnet}
Dongyu Meng and Hao Chen.
\newblock Magnet: a two-pronged defense against adversarial examples.
\newblock In {\em Proceedings of the 2017 ACM SIGSAC conference on computer and
  communications security}, pages 135--147, 2017.

\bibitem{xu2017feature}
Weilin Xu, David Evans, and Yanjun Qi.
\newblock Feature squeezing: Detecting adversarial examples in deep neural
  networks.
\newblock In {\em Network and Distributed System Security Symposium}, 2017.

\bibitem{gu2014towards}
Shixiang Gu and Luca Rigazio.
\newblock Towards deep neural network architectures robust to adversarial
  examples.
\newblock {\em arXiv preprint arXiv:1412.5068}, 2014.

\bibitem{bhagoji2018enhancing}
Arjun~Nitin Bhagoji, Daniel Cullina, Chawin Sitawarin, and Prateek Mittal.
\newblock Enhancing robustness of machine learning systems via data
  transformations.
\newblock In {\em 2018 52nd Annual Conference on Information Sciences and
  Systems (CISS)}, pages 1--5. IEEE, 2018.

\bibitem{Jia_2019_CVPR}
Xiaojun Jia, Xingxing Wei, Xiaochun Cao, and Hassan Foroosh.
\newblock Comdefend: An efficient image compression model to defend adversarial
  examples.
\newblock In {\em Proceedings of the IEEE/CVF Conference on Computer Vision and
  Pattern Recognition (CVPR)}, June 2019.

\bibitem{liu2020defending}
Wenqing Liu, Miaojing Shi, Teddy Furon, and Li~Li.
\newblock Defending adversarial examples via dnn bottleneck reinforcement.
\newblock In {\em Proceedings of the 28th ACM International Conference on
  Multimedia}, pages 1930--1938, 2020.

\bibitem{andriushchenko2020understanding}
Maksym Andriushchenko and Nicolas Flammarion.
\newblock Understanding and improving fast adversarial training.
\newblock {\em Advances in Neural Information Processing Systems}, 33, 2020.

\bibitem{metzen2017detecting}
Jan~Hendrik Metzen, Tim Genewein, Volker Fischer, and Bastian Bischoff.
\newblock On detecting adversarial perturbations.
\newblock {\em arXiv preprint arXiv:1702.04267}, 2017.

\bibitem{grosse2017statistical}
Kathrin Grosse, Praveen Manoharan, Nicolas Papernot, Michael Backes, and
  Patrick McDaniel.
\newblock On the (statistical) detection of adversarial examples.
\newblock {\em arXiv preprint arXiv:1702.06280}, 2017.

\bibitem{hinton2007learning}
Geoffrey~E Hinton.
\newblock Learning multiple layers of representation.
\newblock {\em Trends in cognitive sciences}, 11(10):428--434, 2007.

\bibitem{10.1007/978-981-15-6353-9_34}
Huzaifa~M. Maniyar, Nahid Guard, and Suneeta~V. Budihal.
\newblock Stacked denoising autoencoder: A learning-based algorithm for the
  reconstruction of handwritten digits.
\newblock In Chhabi~Rani Panigrahi, Bibudhendu Pati, Prasant Mohapatra,
  Rajkumar Buyya, and Kuan-Ching Li, editors, {\em Progress in Advanced
  Computing and Intelligent Engineering}, pages 377--387, Singapore, 2021.
  Springer Singapore.

\bibitem{kokalj2019mitigation}
Silvija Kokalj-Filipovic, Rob Miller, Nicholas Chang, and Chi~Leung Lau.
\newblock Mitigation of adversarial examples in rf deep classifiers utilizing
  autoencoder pre-training.
\newblock In {\em 2019 International Conference on Military Communications and
  Information Systems (ICMCIS)}, pages 1--6. IEEE, 2019.

\bibitem{LIU201711}
Weibo Liu, Zidong Wang, Xiaohui Liu, Nianyin Zeng, Yurong Liu, and Fuad~E.
  Alsaadi.
\newblock A survey of deep neural network architectures and their applications.
\newblock {\em Neurocomputing}, 234:11 -- 26, 2017.

\bibitem{shankar2020hyperparameter}
K~Shankar, Yizhuo Zhang, Yiwei Liu, Ling Wu, and Chi-Hua Chen.
\newblock Hyperparameter tuning deep learning for diabetic retinopathy fundus
  image classification.
\newblock {\em IEEE Access}, 8:118164--118173, 2020.

\bibitem{ma2020autonomous}
Benteng Ma, Xiang Li, Yong Xia, and Yanning Zhang.
\newblock Autonomous deep learning: A genetic dcnn designer for image
  classification.
\newblock {\em Neurocomputing}, 379:152--161, 2020.

\bibitem{ozbayoglu2020deep}
Ahmet~Murat Ozbayoglu, Mehmet~Ugur Gudelek, and Omer~Berat Sezer.
\newblock Deep learning for financial applications: A survey.
\newblock {\em Applied Soft Computing}, page 106384, 2020.

\bibitem{bi2020computer}
Xiuli Bi, Shutong Li, Bin Xiao, Yu~Li, Guoyin Wang, and Xu~Ma.
\newblock Computer aided alzheimer's disease diagnosis by an unsupervised deep
  learning technology.
\newblock {\em Neurocomputing}, 392:296--304, 2020.

\bibitem{song2020english}
Zhaojuan Song.
\newblock English speech recognition based on deep learning with multiple
  features.
\newblock {\em Computing}, 102(3):663--682, 2020.

\bibitem{yang2020netflow}
Chao-Tung Yang, Jung-Chun Liu, Endah Kristiani, Ming-Lun Liu, Ilsun You, and
  Giovanni Pau.
\newblock Netflow monitoring and cyberattack detection using deep learning with
  ceph.
\newblock {\em IEEE Access}, 8:7842--7850, 2020.

\bibitem{al2020ensemble}
Abdulrahman Al-Abassi, Hadis Karimipour, Ali Dehghantanha, and Reza~M Parizi.
\newblock An ensemble deep learning-based cyber-attack detection in industrial
  control system.
\newblock {\em IEEE Access}, 8:83965--83973, 2020.

\bibitem{le2013building}
Quoc~V Le.
\newblock Building high-level features using large scale unsupervised learning.
\newblock In {\em 2013 IEEE international conference on acoustics, speech and
  signal processing}, pages 8595--8598. IEEE, 2013.

\bibitem{krizhevsky2017imagenet}
Alex Krizhevsky, Ilya Sutskever, and Geoffrey~E Hinton.
\newblock Imagenet classification with deep convolutional neural networks.
\newblock {\em Communications of the ACM}, 60(6):84--90, 2017.

\bibitem{simonyan2014very}
Karen Simonyan and Andrew Zisserman.
\newblock Very deep convolutional networks for large-scale image recognition.
\newblock {\em Computer ence}, 2014.

\bibitem{szegedy2016inception}
Christian Szegedy, Sergey Ioffe, Vincent Vanhoucke, and Alex Alemi.
\newblock Inception-v4, inception-resnet and the impact of residual connections
  on learning.
\newblock {\em arXiv preprint arXiv:1602.07261}, 2016.

\bibitem{he2015delving}
Kaiming He, Xiangyu Zhang, Shaoqing Ren, and Jian Sun.
\newblock Delving deep into rectifiers: Surpassing human-level performance on
  imagenet classification.
\newblock In {\em Proceedings of the IEEE international conference on computer
  vision}, pages 1026--1034, 2015.

\bibitem{yuan2019adversarial}
Xiaoyong Yuan, Pan He, Qile Zhu, and Xiaolin Li.
\newblock Adversarial examples: Attacks and defenses for deep learning.
\newblock {\em IEEE transactions on neural networks and learning systems},
  30(9):2805--2824, 2019.

\bibitem{8294186}
N.~{Akhtar} and A.~{Mian}.
\newblock Threat of adversarial attacks on deep learning in computer vision: A
  survey.
\newblock {\em IEEE Access}, 6:14410--14430, 2018.

\bibitem{papernot2016distillation}
Nicolas Papernot, Patrick McDaniel, Xi~Wu, Somesh Jha, and Ananthram Swami.
\newblock Distillation as a defense to adversarial perturbations against deep
  neural networks.
\newblock In {\em 2016 IEEE symposium on security and privacy (SP)}, pages
  582--597. IEEE, 2016.

\bibitem{samangouei2018defense}
Pouya Samangouei, Maya Kabkab, and Rama Chellappa.
\newblock Defense-gan: Protecting classifiers against adversarial attacks using
  generative models.
\newblock {\em arXiv preprint arXiv:1805.06605}, 2018.

\bibitem{song2017pixeldefend}
Yang Song, Taesup Kim, Sebastian Nowozin, Stefano Ermon, and Nate Kushman.
\newblock Pixeldefend: Leveraging generative models to understand and defend
  against adversarial examples.
\newblock {\em arXiv preprint arXiv:1710.10766}, 2017.

\bibitem{carlini2017adversarial}
Nicholas Carlini and David Wagner.
\newblock Adversarial examples are not easily detected: Bypassing ten detection
  methods.
\newblock In {\em Proceedings of the 10th ACM Workshop on Artificial
  Intelligence and Security}, pages 3--14, 2017.

\bibitem{athalye2018obfuscated}
Anish Athalye, Nicholas Carlini, and David Wagner.
\newblock Obfuscated gradients give a false sense of security: Circumventing
  defenses to adversarial examples.
\newblock In {\em International Conference on Machine Learning}, pages
  274--283. PMLR, 2018.

\bibitem{biggio2013evasion}
Battista Biggio, Igino Corona, Davide Maiorca, Blaine Nelson, Nedim
  {\v{S}}rndi{\'c}, Pavel Laskov, Giorgio Giacinto, and Fabio Roli.
\newblock Evasion attacks against machine learning at test time.
\newblock In {\em Joint European conference on machine learning and knowledge
  discovery in databases}, pages 387--402. Springer, 2013.

\bibitem{8482346}
B.~{Liang}, H.~{Li}, M.~{Su}, X.~{Li}, W.~{Shi}, and X.~{Wang}.
\newblock Detecting adversarial image examples in deep neural networks with
  adaptive noise reduction.
\newblock {\em IEEE Transactions on Dependable and Secure Computing},
  18(1):72--85, 2021.

\bibitem{lecun1998mnist}
Yann LeCun.
\newblock The mnist database of handwritten digits.
\newblock {\em http://yann. lecun. com/exdb/mnist/}, 1998.

\bibitem{krizhevsky2009learning}
Alex Krizhevsky, Geoffrey Hinton, et~al.
\newblock Learning multiple layers of features from tiny images.
\newblock 2009.

\bibitem{10.1145/1390156.1390294}
Pascal Vincent, Hugo Larochelle, Yoshua Bengio, and Pierre-Antoine Manzagol.
\newblock Extracting and composing robust features with denoising autoencoders.
\newblock In {\em Proceedings of the 25th International Conference on Machine
  Learning}, ICML '08, page 1096–1103, New York, NY, USA, 2008. Association
  for Computing Machinery.

\bibitem{BELLO2020259}
Marilyn Bello, Gonzalo Nápoles, Ricardo Sánchez, Rafael Bello, and Koen
  Vanhoof.
\newblock Deep neural network to extract high-level features and labels in
  multi-label classification problems.
\newblock {\em Neurocomputing}, 413:259 -- 270, 2020.

\bibitem{10.1007/978-3-030-50516-5_13}
Qiang Zhao and Fakhri Karray.
\newblock Anomaly detection for images using auto-encoder based sparse
  representation.
\newblock In Aur{\'e}lio Campilho, Fakhri Karray, and Zhou Wang, editors, {\em
  Image Analysis and Recognition}, pages 144--153, Cham, 2020. Springer
  International Publishing.

\bibitem{9248108}
F.~{Zhang} and H.~{Fleyeh}.
\newblock Anomaly detection of heat energy usage in district heating
  substations using lstm based variational autoencoder combined with physical
  model.
\newblock In {\em 2020 15th IEEE Conference on Industrial Electronics and
  Applications (ICIEA)}, pages 153--158, 2020.

\bibitem{9113298}
S.~{Zavrak} and M.~{İskefiyeli}.
\newblock Anomaly-based intrusion detection from network flow features using
  variational autoencoder.
\newblock {\em IEEE Access}, 8:108346--108358, 2020.

\bibitem{2016cleverhans}
Nicolas Papernot, Ian Goodfellow, Ryan Sheatsley, Reuben Feinman, and Patrick
  Mcdaniel.
\newblock cleverhans v1.0.0: an adversarial machine learning library.
\newblock 2016.

\end{thebibliography}

%

\begin{IEEEbiography}[{\includegraphics[width=1in,height=1.25in,clip,keepaspectratio]{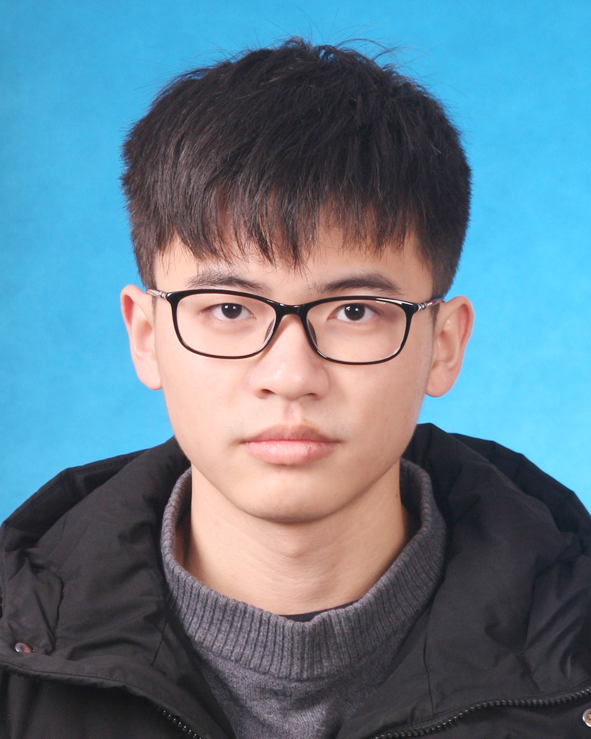}}]{Haowen Liu} is a senior undergraduate student at School of Electronic Information and Electrical Engineering, Shanghai Jiao Tong University in China. He is currently studying at School of Cyber Science and Engineering. His research interests include computer security, cyber security, artificial intelligence security, and augmented reality.
\end{IEEEbiography}

\begin{IEEEbiography}[{\includegraphics[width=1in,height=1.25in,clip,keepaspectratio]{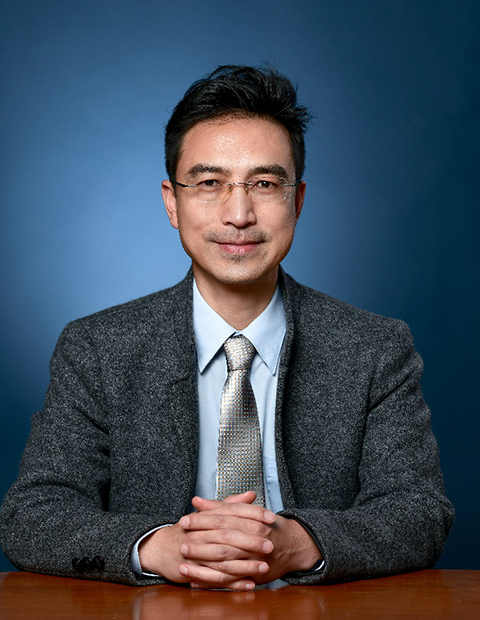}}]{Ping Yi} is an Associate Professor at School of Electronic Information and Electrical Engineering, Shanghai Jiao Tong University in China. He received the Ph.D degree at the department of Computing and Information Technology, Fudan University, China. His research interests include artificial intelligence security. He is a member of IEEE Communications and Information Security Technical Committee. He is a senior member of IEEE.
\end{IEEEbiography}


\begin{IEEEbiography}[{\includegraphics[width=1in,height=1.25in,clip,keepaspectratio]{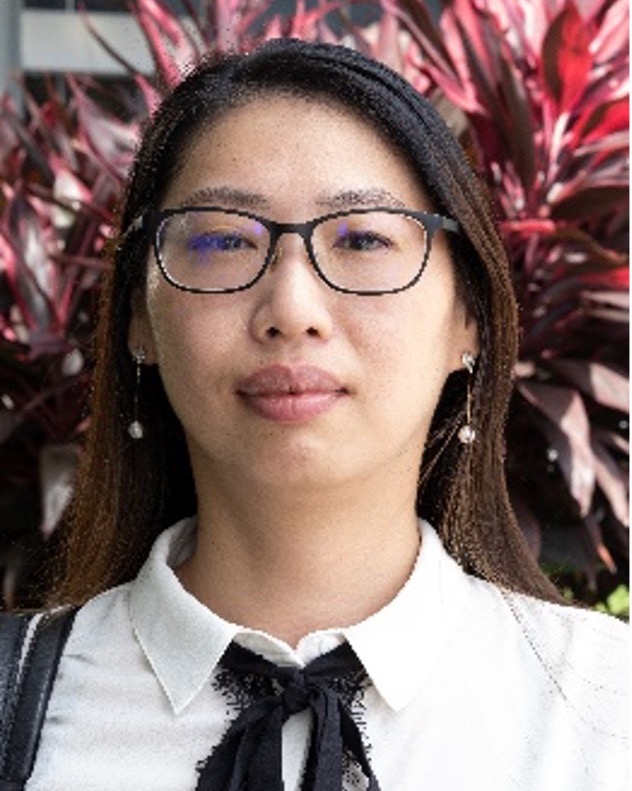}}]{Hsiao-Ying Lin} is a senior researcher in Shield Labs at Huawei International. Her research interests include adversarial machine learning, applied cryptography and security issues in automotive areas. She received the MS and PhD degrees in computer science from National Chiao Tung University, Taiwan, in 2005 and 2010, respectively. She is an IEEE member.
\end{IEEEbiography}

\begin{IEEEbiography}[{\includegraphics[width=1in,height=1.25in,clip,keepaspectratio]{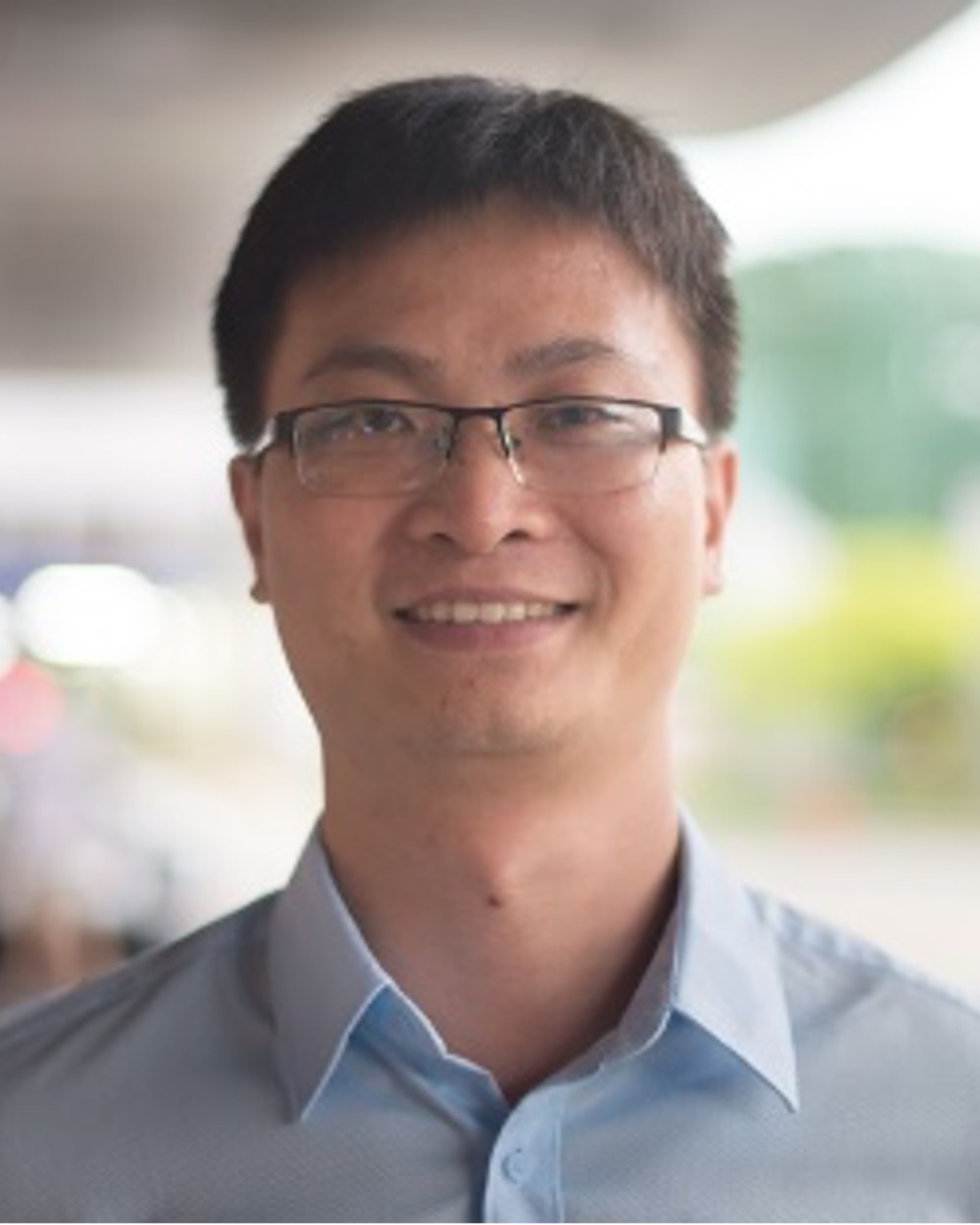}}]{Jie Shi} is a Security Expert in Huawei Singpaore Research Center. His research interests include trustworthy AI, machine learning security, data security and applied cryptography. He received his Ph.D degree from Huazhong University of Science and Technology, China.
\end{IEEEbiography}

\begin{IEEEbiography}[{\includegraphics[width=1in,height=1.25in,clip,keepaspectratio]{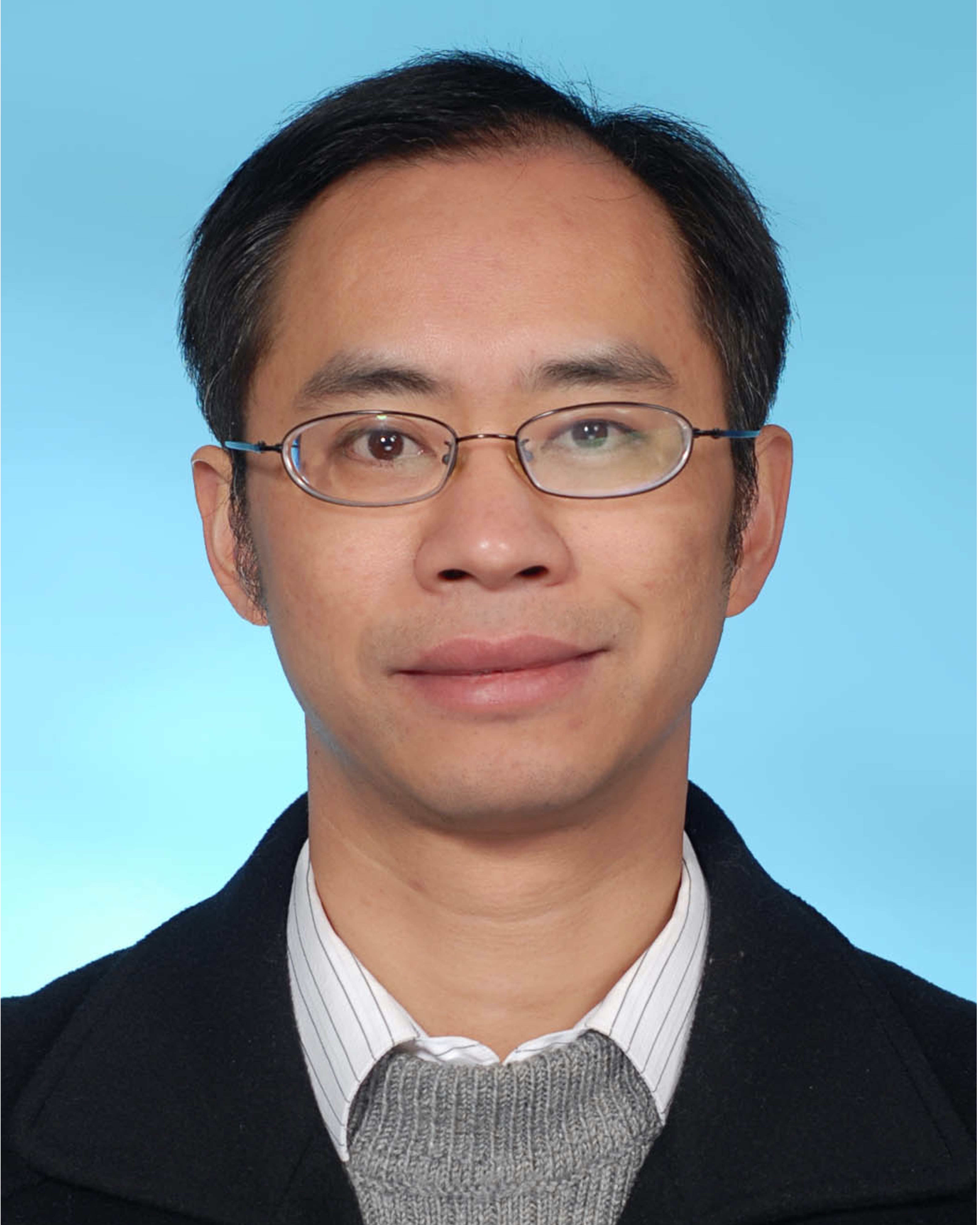}}]{Weidong Qiu} received the M.S. degree in cryptography from Xidian University, Xi’an, China, in 1998 and Ph.D. degree in computer software theory from Shanghai Jiao Tong University, Shanghai, China, in 2001. He is currently a Professor in the School of Cyber Security, Shanghai Jiao Tong University.  He has published more than ninety academic papers on cryptology. His main research areas include cryptography and computer forensics. He is a senior member of IEEE.
\end{IEEEbiography}




\end{document}